\documentclass[10pt,journal,compsoc]{IEEEtran}

\usepackage{amsmath,amsfonts}
\usepackage{algorithmic}
\usepackage{algorithm}
\usepackage{array}
\usepackage[caption=false,font=normalsize,labelfont=sf,textfont=sf]{subfig}
\usepackage{textcomp}
\usepackage{stfloats}
\usepackage{url}
\usepackage{verbatim}
\usepackage{graphicx}
\usepackage{cite}
\usepackage{multirow}
\usepackage{pifont}

\usepackage[pagebackref=true,breaklinks=true,colorlinks,bookmarks=false]{hyperref}

\begin{document}

\title{Planner3D: LLM-enhanced graph prior meets 3D indoor scene explicit regularization}

\author{Yao Wei, Martin Renqiang Min, George Vosselman, Li Erran Li, Michael Ying Yang

\IEEEcompsocitemizethanks{
\IEEEcompsocthanksitem Yao Wei and George Vosselman are with University of Twente, The Netherlands.
\IEEEcompsocthanksitem Martin Renqiang Min is with NEC Lab America, USA.
\IEEEcompsocthanksitem Li Erran Li is with AWS AI, Amazon, USA.
\IEEEcompsocthanksitem  Michael Ying Yang is with Viusal Computing Group, University of Bath, UK. Email:
\url{myy35@bath.ac.uk}.
\IEEEcompsocthanksitem Project webpage: \href{https://sites.google.com/view/planner3d/}{https://sites.google.com/view/Planner3D/}
}

}

\IEEEtitleabstractindextext{%
\begin{abstract}
Compositional 3D scene synthesis has diverse applications across a spectrum of industries such as robotics, films, and video games, as it closely mirrors the complexity of real-world multi-object environments. Conventional works typically employ shape retrieval based frameworks which naturally suffer from limited shape diversity. Recent progresses have been made in object shape generation with generative models such as diffusion models, which increases the shape fidelity. However, these approaches separately treat 3D shape generation and layout generation. The synthesized scenes are usually hampered by layout collision, which suggests that the scene-level fidelity is still under-explored. In this paper, we aim at generating realistic and reasonable 3D indoor scenes from scene graph. To enrich the priors of the given scene graph inputs, large language model is utilized to aggregate the global-wise features with local node-wise and edge-wise features. With a unified graph encoder, graph features are extracted to guide joint layout-shape generation. Additional regularization is introduced to explicitly constrain the produced 3D layouts. Benchmarked on the SG-FRONT dataset, our method achieves better 3D scene synthesis, especially in terms of scene-level fidelity. The source code will be released after publication.
\end{abstract}

}

\maketitle

\IEEEdisplaynontitleabstractindextext

\IEEEpeerreviewmaketitle

\IEEEraisesectionheading{\section{Introduction}\label{sec:introduction}}
\IEEEPARstart{G}{enerative} models \cite{kingma2013auto, goodfellow2014generative, rezende2015variational, ho2020denoising} have shown great potential in computer vision, natural language processing and multi-modal learning. With the learned patterns from generative models, data can be synthesized from scratch or other modalities. For controllable content creation, which has attracted increasing attention from both academia and industry, a variety of conditional generative methods have emerged on tasks such as text-to-image \cite{ramesh2022hierarchical, tang2023dreamgaussian}, image-to-3D \cite{liu2023zero, liu2024one} and text-to-3D \cite{jun2023shap, tian2023shapescaffolder} generation. These cross-modal synthesis methods significantly reduce the data acquisition budget, especially for 3D content whose collection requires a lot of human efforts.

Despite the recent impressive breakthroughs in 3D content creation, most of the existing works \cite{zhou20213d, liu2022iss, wei2022flow, cheng2023sdfusion} are restricted to single-object scenes, and their performance inevitably degrades on compositional scenes due to the limited understanding of object-object and scene-object relationships. Even trained on Internet-scale captioned 3D objects, for instance, Shap-E \cite{jun2023shap} struggles to reliably produce the results that are consistent with the given spatial or numerical prompts. Moreover, labor-intensive post-processing is essential to scale these approaches to complex scenes containing multiple objects, since the synthesized 3D shapes are sometimes in the non-canonical space. 

\begin{figure*}[tb]
  \centering
  \includegraphics[width=.98\textwidth]{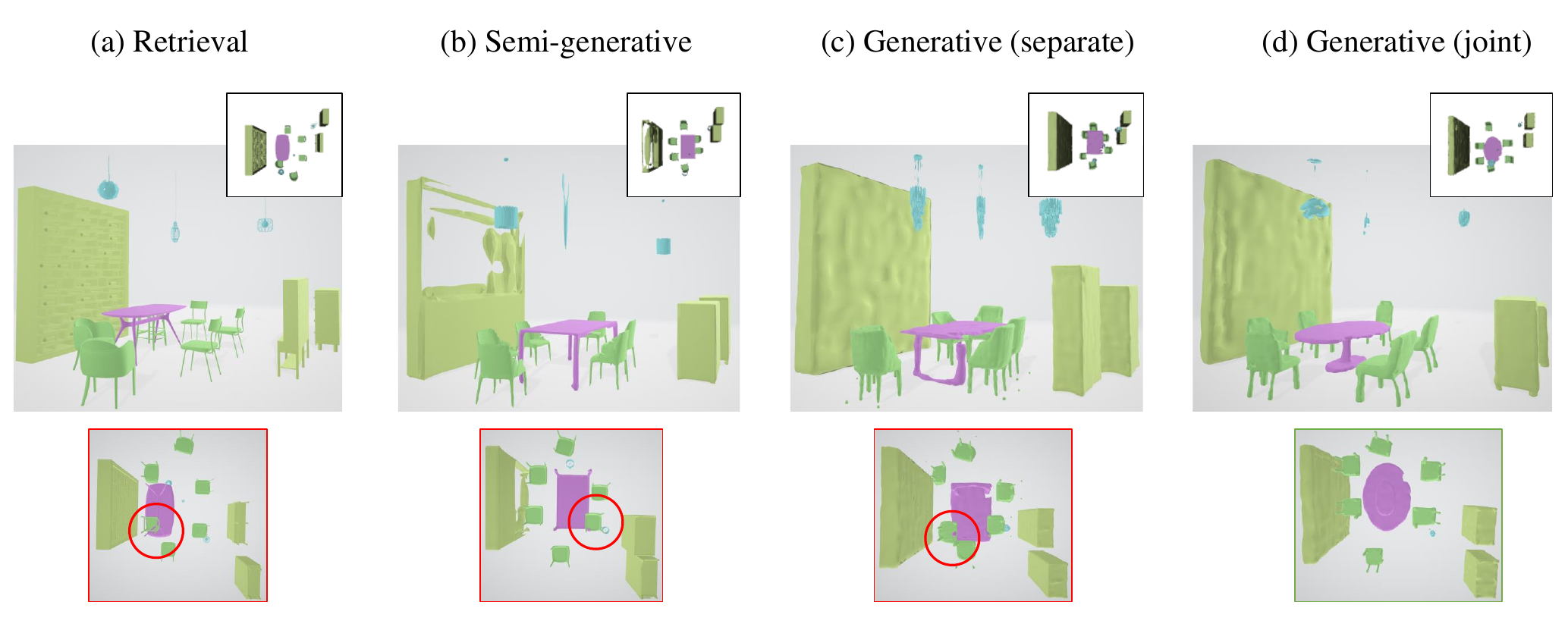}
  \caption{Overview of previous works and ours. The 3D scenes are rendered from top-down view (top right), side view (middle), and bottom-up view (bottom). Unlike previous works that show poor shape consistency or spatial arrangements, Planner3D can synthesize higher-fidelity 3D scenes which demonstrate more realistic layout configuration while preserving shape consistency and diversity. 
  }
  \label{fig:overview}
\end{figure*}

In recent years, new pipelines capable of producing the entire 3D scenes have been developed using diverse inputs, such as LiDAR scans \cite{ren2023xcube}, single RGB images \cite{engelmann2021points}, captioned bounding boxes (i.e., scene layouts) \cite{po2023compositional} and scene graphs \cite{luo2020end, dhamo2021graph, zhai2023commonscenes, zhai2024echoscene}. As a semantic representation of 3D scenes, scene graphs provide a promising interface to control the generated 3D content, e.g., assigning properties to different instances of the same object category. However, 3D scene synthesis from scene graph is still at an early stage, and existing methods encounter various challenges.

Conventional works focus on scene layout generation regardless of 3D shape generation. Instead, the object shapes are retrieved from the existing shape databases. Though benefiting from reliable shape quality, as illustrated in Fig. \ref{fig:overview} (a), it is hard for retrieval-based methods to generate coherent 3D scenes. For example, chairs appear in arbitrary shape styles. In addition, the performance heavily depends on the scale of the given databases. Differently, recent works synthesize the 3D scenes from generative perspectives. By category-specified shape decoders, semi-generative method Graph-to-3D \cite{dhamo2021graph} alleviates the limitation of retrieval-based method on shape consistency, but it suffers from low-quality shapes such as the incomplete wine cabinet shown in Fig. \ref{fig:overview} (b). Fully-generative method CommonScenes \cite{zhai2023commonscenes} enables generative shape synthesis by adopting a latent diffusion model, which enhances the shape fidelity and diversity. Nevertheless, it can be seen from Fig. \ref{fig:overview} (c) that the generated objects are spatially ambiguous due to the generated low-quality scene layouts. In fact, this frequently appears in (a), (b) and (c) where the table intersects the adjacent chair, denoted with red circles in the bottom-up views. These unrealistic scene generations leave a huge gap between previous methods and practical applications.

For controllable 3D scene synthesis, the generated objects should be not only realistic on its own but in harmony with the context of scene. To this end, we present \textbf{Planner3D}, \textbf{L}LM-enhanced gr\textbf{A}ph \textbf{P}rior meets \textbf{3D} i\textbf{N}door sce\textbf{N}e \textbf{E}xplicit \textbf{R}egularization, for scene graph guided 3D scene synthesis. Instead of treating the compositional scene synthesis as two separate tasks, we work on joint layout-shape generation and explore controllable generative models by collaborating LLMs. Specifically, LLMs enhance the scene graph representation via interpreting the whole graph, which enables hierarchical understanding of the graph priors. Driven by a unified graph encoder, shape branch and layout branch are jointly trained to synthesize 3D scenes. Moreover, a layout regularization loss is introduced to explicitly encourage reasonable arrangements. As illustrated in Fig. \ref{fig:overview} (d), our method is capable of generating realistic compositional 3D scenes. The \textbf{contributions} made in this work can be summarized as follows. 
\begin{itemize}
\item We propose Planner3D, a controllable generative pipeline for compositional 3D indoor scene synthesis given scene graph. Planner3D effectively resolves the issues of unrealistic scenes and object collisions that plagued previous works.
\item Scene graph prior enhancement mechanism is designed by adopting LLMs to produce global-wise textual representation from the graph, leading to hierarchical graph priors.
\item A layout regularization loss, which explicitly constraints spatial arrangements and prevents unexpected intersections between objects, is introduced to a dual-branch graph-to-scene generation framework.
\item Experimental results demonstrate that Planner3D improves the fidelity of scene synthesis. It attains 2.55 lower Fréchet Inception Distance (FID) score and 1.24 lower Kernel Inception Distance (KID) score compared to SOTA CommonScenes \cite{zhai2023commonscenes}.
\end{itemize}


\section{Related Work}

In this section, we discuss the most relevant literature on generative models for content creation, especially compositional 3D scene synthesis. We also survey emerging Large Language Models (LLMs) which are relevant to our work.

\subsection{Generative models for content creation}
Generative models have gained wide attention in recent years, with several prominent architectures emerging, including Variational Autoencoders (VAEs) \cite{kingma2013auto}, Generative Adversarial Networks (GANs) \cite{goodfellow2014generative}, Normalizing Flows (Flows) \cite{rezende2015variational}, and Denoising Diffusion Probabilistic Models (DDPMs) \cite{ho2020denoising}. With the rapid improvements of deep learning techniques, these models are capable of creating novel content based on the learned distribution of training data. Using conditioning information, generative models can also be applied to personalized content creation, i.e., conditional generation. For example, taking natural language descriptions as the conditioning text prompts, there are many cross-modal applications, e.g., text-to-speech \cite{kim2021conditional,tan2024naturalspeech}, text-to-image \cite{ramesh2022hierarchical, tang2023dreamgaussian}, text-to-video \cite{ma2024follow, wang2024motionctrl}, and text-to-3D \cite{jun2023shap, tian2023shapescaffolder} synthesis. While remarkable progress has been made, challenges remain in conditional generation, particularly with regard to the controllability of generative models. Additionally, conditional generation lays behind in modalities such as 3D domain, due to the lack of large-scale datasets with high-quality condition-3D data pairs, and the complexity of modeling high-dimensional data.

For conditional 3D generation using generative models, most of the existing studies \cite{zhou20213d, liu2022iss, wei2022flow, cheng2023sdfusion, tian2023shapescaffolder, jun2023shap} primarily focus on synthesizing 3D scenes containing a single object. The object shapes are typically represented by explicit 3D representations that directly define the geometry of objects, such as voxel grids \cite{hane2017hierarchical}, point clouds \cite{achlioptas2018learning}, and meshes \cite{wang20193dn}. Multi-view images \cite{su2015multi} are commonly used to capture the object's appearance from various angles. Alternatively, some works employ implicit representations such as Signed Distance Functions (SDFs) \cite{chen2019learning} which model the 3D surfaces through continuous functions, and Neural Radiance Fields (NeRFs) \cite{mildenhall2020nerf} which model the volumetric scenes using neural networks. With the recent advances of 3D Gaussian Splatting \cite{kerbl20233d}, the hybrid explicit-and-implicit representation \cite{zou2024triplane} is also explored. In this work, we use SDFs for geometry representation which allows us to model multi-object scenes in a compositional manner.

\subsection{Compositional 3D scene synthesis}

Unlike object-centric 3D synthesis, multi-object 3D scene synthesis is more challenging where the object attributes (e.g., categories) and relationships become both crucial. Prior works \cite{luo2020end, engelmann2021points, po2023compositional, ren2023xcube, dhamo2021graph, zhai2023commonscenes} usually characterize a scene by spatial arrangements (i.e., layouts) and 3D shapes of the objects-of-interest within the scene. According to the styles of shape synthesis, these works can be broadly divided into three types of approaches.

\vspace{2mm}
\noindent\textbf{Retrieval based approaches.}
For simplicity, conventional works \cite{luo2020end, engelmann2021points, paschalidou2021atiss, tang2024diffuscene, lin2024instructscene} formulate the task as a combination of 3D layout prediction problem and 3D shape retrieval problem. They mainly concentrate on the former problem by training generative models to estimate 3D layouts given the input condition, such as scene graph \cite{luo2020end}, a single RGB image \cite{engelmann2021points}, and text prompts \cite{tang2024diffuscene, lin2024instructscene}. The layouts are then populated with category-aware shapes to construct the 3D scenes. Specifically, the shapes are retrieved from the open-sourced databases, such as ShapeNet \cite{chang2015shapenet}, 3D-FUTURE \cite{fu20213d2} and Objaverse \cite{deitke2023objaverse, deitke2024objaverse}, based on euclidean distance of the bounding box dimensions \cite{paschalidou2021atiss}. However, this formulation depends heavily on the scale of the databases, suffering from limited shape diversity as well as inconsistent shapes of the same object category. 

\vspace{2mm}
\noindent\textbf{Semi-generative approaches.}
Concurrent studies \cite{ren2023xcube, po2023compositional, bahmani2023cc3d} focus on generative shape synthesis, while scarifying the layout synthesis. Conditioned on LiDAR point clouds, XCube \cite{ren2023xcube} can produce 3D scenes with high-quality meshes, though these partial observations are apparently not user-friendly inputs. Given axis-aligned 3D layouts that roughly provide the location and size of objects, Comp3D \cite{po2023compositional} is able to synthesize compositional 3D scenes that fit the specified object categories. CC3D \cite{bahmani2023cc3d} generates 3D scenes from 2D semantic layouts. While benefiting from extensive image data and 3D field representations, the results of layout-conditioned scene synthesis often exhibit view inconsistency problem. Also, these approaches impose strong geometric priors, significantly limiting the diversity of generated scenes. Graph-to-3D \cite{dhamo2021graph}, a pioneering work in jointly shape and layout synthesis from scene graphs, overcomes the limitations of previous works in consistency and diversity. Nevertheless, its shape synthesis is bounded by pre-trained category-wise shape decoders.

\vspace{2mm}
\noindent\textbf{Fully-generative approaches.} Leveraging latent diffusion-based shape synthesis, CommonScenes \cite{zhai2023commonscenes} presents a fully-generative pipeline for 3D scene synthesis from scene graphs, achieving high-fidelity generation compared to previous studies. However, it suffers from three major issues. First, CommonScenes attempts to implicitly enrich scene graphs with contextual information which is restricted to individual edge-wise textual representations. The absence of global-wise explicit graph priors hinders effective feature extraction by graph convolution networks. Second, it struggles with generating coherent 3D scenes since the shape synthesis is isolated from the layout synthesis. Moreover, it mostly relies on adversarial training strategy for layout synthesis, resulting in unrealistic layouts and object collisions. Motivated by addressing these problems, we involves LLMs for reasoning about the object-object and object-scene relationships from scene graphs. We also revisit the architecture and training scheme of 3D scene synthesis, especially in terms of graph encoder and layout decoder.

\subsection{Annotations generated by LLMs}

High quality labels play a vital role for machine learning (ML) systems.
Traditionally, they are collected from humans manually or with limited ML assistance. Due to the high cost, large datasets with high quality text annotations are rare which hampers progress in the field. As LLMs are very capable, prior works have successfully employed LLMs to automatically generate annotations. In the NLP domain, besides earlier successes on generating annotations for tasks such as relevance detection, topic classification, and sentiment analysis, recently~\cite{Orca2023} shows GPT-4 can even generate data to solve advanced reasoning tasks. In the realm of images, LLMs are widely employed to generate question-answer text pairs from image captions for instruction tuning~\cite{LLaVA2023, ViPLLaVA2023, MiniGPT2023}. In video datasets, LLMs are primarily employed to enrich the text descriptions of videos. For instance, MAXI~\cite{MAXI2022} leverages LLMs to generate additional textual descriptions of videos from their category names for training. InternVideo~\cite{InternVideo2022} utilizes LLMs to generate the action-level descriptions of videos from their frame-level captions. HowtoCap~\cite{HowtoCap2023} leverages LLMs to enhance the detail in narrations of instructional videos. Lately, GraphDreamer \cite{gao2024graphdreamer} generates 3D scenes from texts by adpoting LLMs to produce scene graphs as intermediate representations. LayoutGPT \cite{feng2024layoutgpt} regards LLMs as text-driven layout planners for scene synthesis. By contrast, we employ LLMs to generate global-wise textual descriptions of the scene graph. These rich and consistent descriptions prove to be vital to improve the representation learning in order to generate realistic compositional 3D scenes. 

\begin{figure*}[tb]
  \centering
  \includegraphics[width=.98\textwidth]{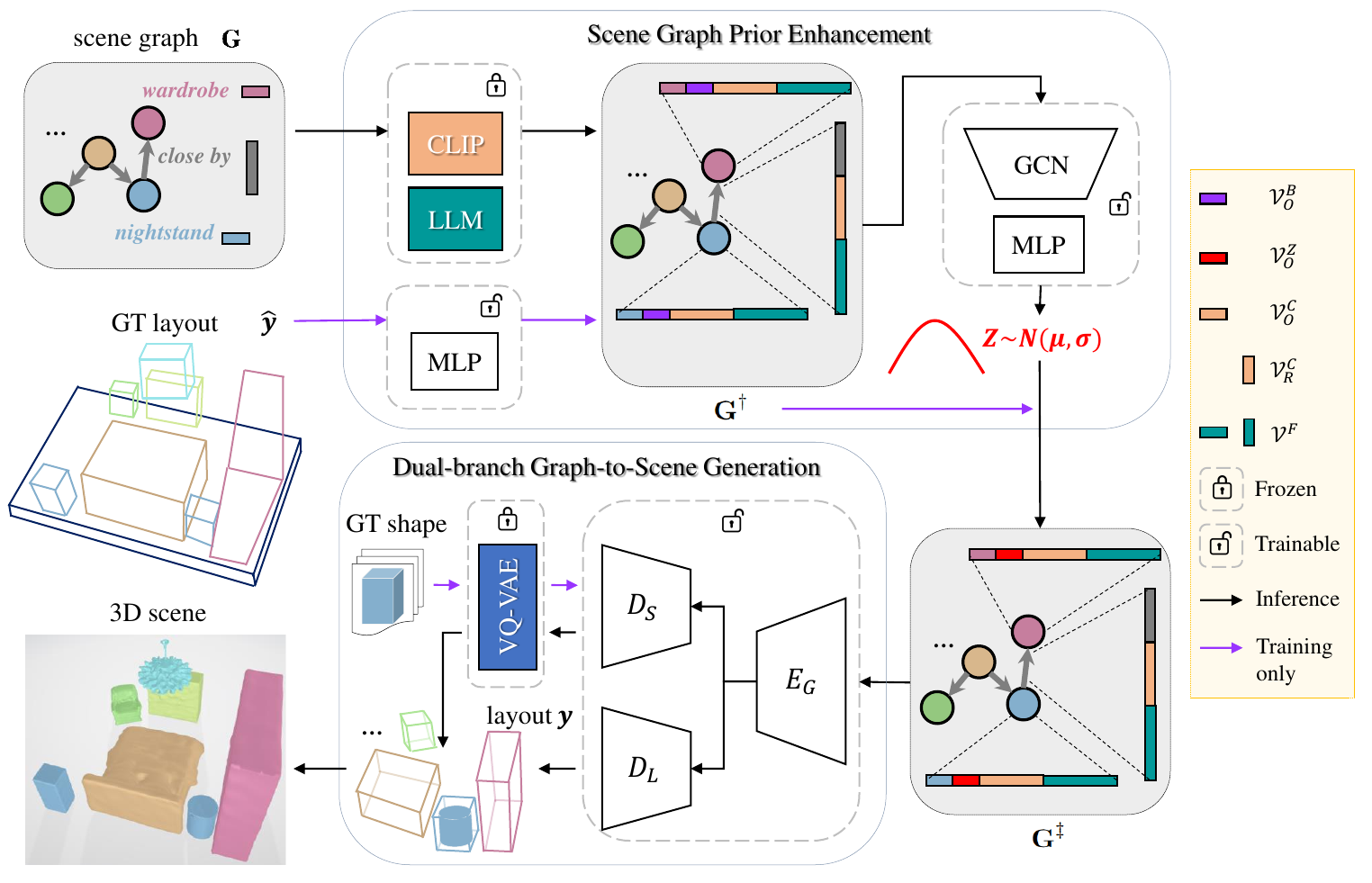}
  \caption{Overview of Planner3D. Given a scene graph $\textbf{G}$, graph prior is enriched with LLM and CLIP to construct graph representation $\textbf{G}^\dag$, which is then used to model the distribution $Z$. After updating graph node representation by replacing layout vectors with random vectors $z$ sampled from $Z$, the updated graph representation $\textbf{G}^\ddag$ is input into the graph encoder $E_G$. Guided by the extracted graph features, consequently, the layout decoder $D_L$ generates 7-parameterized layouts, and the shape decoder $D_S$ works in conjunction with VQ-VAE to synthesize object shapes for graph nodes.   
  }
  \label{fig:pipeline}
\end{figure*}

\section{Method}

Given a scene graph $\textbf{G}$ describing the desired multi-object 3D scene using objects as nodes and their relationships as edges, Planner3D is able to synthesize realistic 3D scenes with consistent 3D object shapes and spatial layouts. As illustrated in Fig. \ref{fig:pipeline}, Planner3D consists of two main components: a scene graph prior enhancement mechanism and a dual-branch encoder-decoder architecture for graph-to-scene generation. 

First, scene graph prior is enriched with LLM and vision-language model CLIP, and explicitly aggregates node-wise, edge-wise and global-wise textual representations of the input scene graph. The aggregated representation $\textbf{G}^\dag$ is fed to Graph Convolutional Networks (GCNs) and Multilayer Perceptron (MLP) layers, which are trained to model the posterior distribution of the 3D scene conditioned on the given scene graph. With the learned distribution $Z$, we update the node representation by replacing the original layout vector with the random vector $z$ sampled from $Z$. Then, a graph encoder $E_G$ learns to extract graph features from the updated graph representation $\textbf{G}^\ddag$. A scene decoder takes graph features as inputs and learns to generate 7-degrees-of-freedom 3D layout and shape latent through layout decoder $D_L$ and shape decoder $D_S$, respectively. Compositional 3D scenes are eventually synthesized by fitting the generated layouts with the 3D shapes reconstructed from the shape latent by the pre-trained Vector Quantized Variational Autoencoder (VQ-VAE) decoder.

\begin{figure*}[tb]
  \centering
  \includegraphics[width=.98\textwidth]{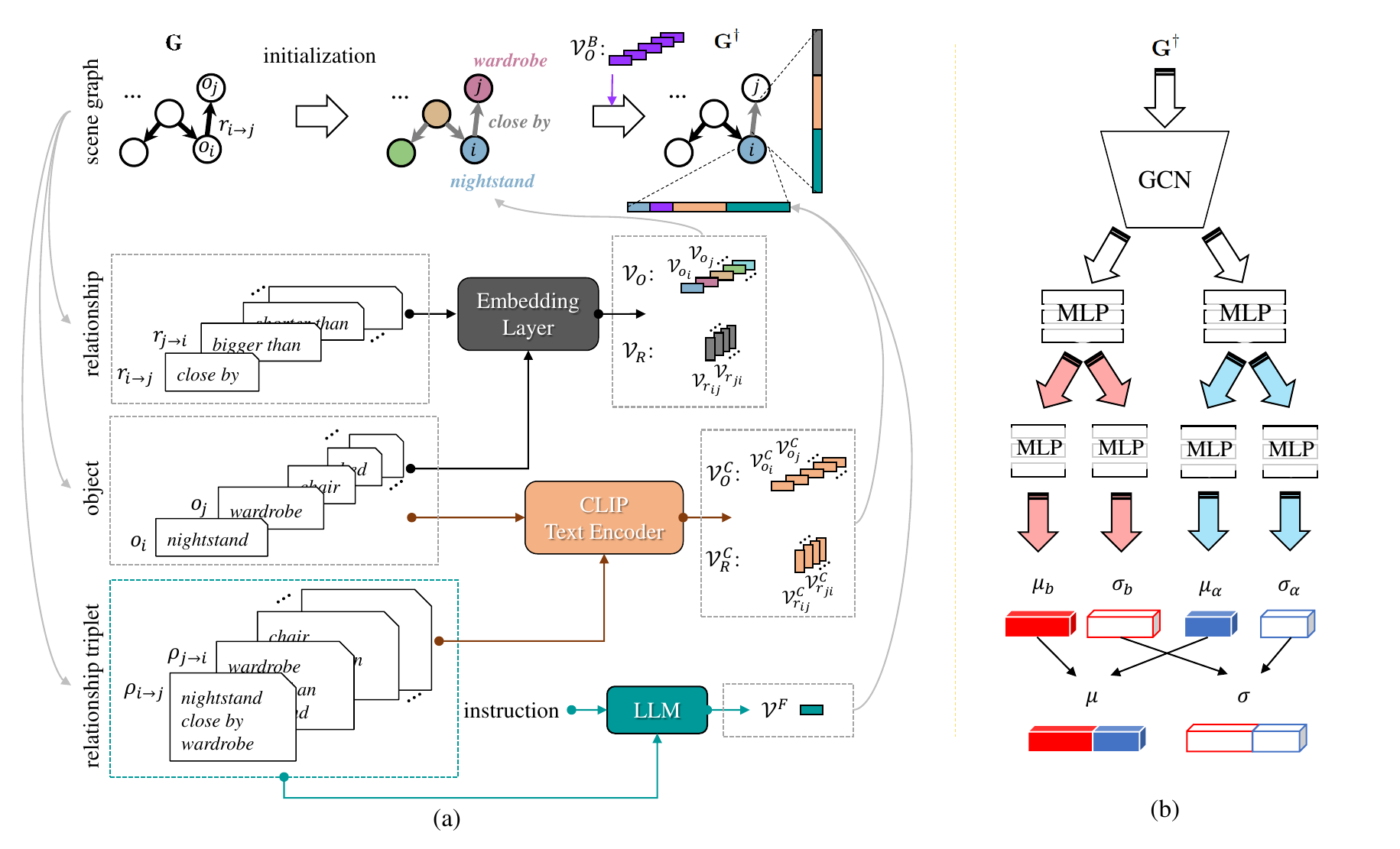}
  \caption{Pipeline of scene graph prior enhancement. (a) Using CLIP and LLM, the graph representation $\textbf{G}^\dag$ explicitly aggregates node-wise, edge-wise and global-wise textual representations of the given scene graph. (b) The architecture of the model $\phi$, which learns to model the distribution $Z\sim\mathcal{N} (\mu,\sigma)$ and is used to construct the graph representation $\textbf{G}^\ddag$.}
  \label{fig:graph}
\end{figure*}

\subsection{Scene Graph Prior Enhancement}

The scene graph \textbf{G} $=(O,R)$ typically consists of $N$ nodes, i.e., objects $O=\{o_1, o_2, ..., o_N\}$, and $M$ directed edges, i.e., object relationships $R=\{r_1, r_2, ..., r_M\}$. The representation of scene graph is initially composed of a set of feature vectors $\mathcal{V}_O \in \mathbb{R}^{\mathrm{N\times Q}}$ and $\mathcal{V}_R \in \mathbb{R}^{\mathrm{M\times 2Q}}$ for nodes and edges, which are embedded from the classes of objects $O$ and relationships $R$, respectively. $Q$ indicates the dimensions of feature vectors. Additionally, layout feature vectors $\mathcal{V}_O^B \in \mathbb{R}^{\mathrm{N\times Q}}$ are embedded using ground-truth (GT) layouts $\hat{y}$ parameterized by 3D bounding box locations and sizes $\hat{b}$ as well as angular rotations $\hat{\alpha}$ along the vertical axis.

As demonstrated in Fig. \ref{fig:graph}(a), the pre-trained and frozen CLIP \cite{radford2021learning} is adopted to enrich the initial graph representation with inter-object insights. Specifically, nodes and edges are regarded as text prompts. Taking the objects $o_i$ and $o_j$ and their relationship $r_{i\rightarrow j}$ as examples, we denote the corresponding relationship triplet \textless subject, predicate, object\textgreater\ as $\rho_{i\rightarrow j}$, where $o_i$, $r_{i\rightarrow j}$ and $o_j$ act as the subject, predicate, and object, respectively. CLIP text encoder is used to derive contextual feature vectors $\mathcal{V}_{r_{ij}}^C\in\mathcal{V}_R^C$ for $\rho_{i\rightarrow j}$. Similarly, the vectors $\mathcal{V}_{o_i}^C,\mathcal{V}_{o_j}^C\in\mathcal{V}_O^C$ are produced with prompts on subjects or objects of relationship triplets.

Furthermore, a pre-trained and frozen LLM is employed to generate feature vectors $\mathcal{V}^F$ for the whole graph, enhancing the expressiveness of scene graph priors. By treating the scene graph as a set of relationship triplets $\Omega=\{\rho_1, \rho_2, ..., \rho_M\}$, Instructor Transformer \cite{su2022one} which serves as LLM in this work, takes aggregated relationship triplets as input prompts and generates text embedding using an instruction with a format of "Represent the \{DOMAIN\} \{TEXT-TYPE\} for \{TASK\}". Here, we specify the instruction by Science, document and summarization, respectively. The resulting LLM-based vector $\mathcal{V}^F$ is attached to each node and edge feature vectors. 

Thanks to CLIP and LLM, the graph representation $\textbf{G}^\dag$ explicitly aggregates scene graph priors at the node-wise, edge-wise and global-wise levels. It can be formulated as,
\begin{equation}
\textbf{G}^\dag = \{\mathcal{V}_O \circ \mathcal{V}_O^B \circ \mathcal{V}_O^C\circ \mathcal{V}^{F},\mathcal{V}_R\circ \mathcal{V}_R^C\circ \mathcal{V}^{F}\},
\label{eq:aggr_graph}
\end{equation}
where $\circ$ indicates the concatenation operation. Then, GCN layers are exploited to pass information along graph edges, followed by two MLP modules in parallel, each of which is composed of a MLP layer and additional two MLP heads. Fig. \ref{fig:graph}(b) shows the architecture of GCNs and MLPs, parameterized as $\phi$. The MLP heads separately output the means (demoted as $\mu_b$ and $\mu_\alpha$) and the variances (denoted as $\sigma_b$ and $\sigma_\alpha$). By concatenating $\mu_b$ and $\mu_\alpha$ as well as $\sigma_b$ and $\sigma_\alpha$, $\mu$ and $\sigma$ are ultimately produced. Through re-parameterization, random vectors $z$ are sampled from the distribution $Z \sim\mathcal{N} (\mu,\sigma)$ for each node. These random vectors $\mathcal{V}_O^Z \in \mathbb{R}^{\mathrm{N\times Q}}$ are utilized to replace the layout vectors $\mathcal{V}_O^B$ that only work for the training phase. The updated graph is represented as follows, 
\begin{equation}
\textbf{G}^\ddag = \{\mathcal{V}_O \circ \mathcal{V}_O^Z \circ \mathcal{V}_O^C\circ \mathcal{V}^{F},\mathcal{V}_R\circ \mathcal{V}_R^C\circ \mathcal{V}^{F}\}.
\label{eq:update_graph}
\end{equation}

\begin{figure}[tb]
  \centering
  \includegraphics[width=\columnwidth]{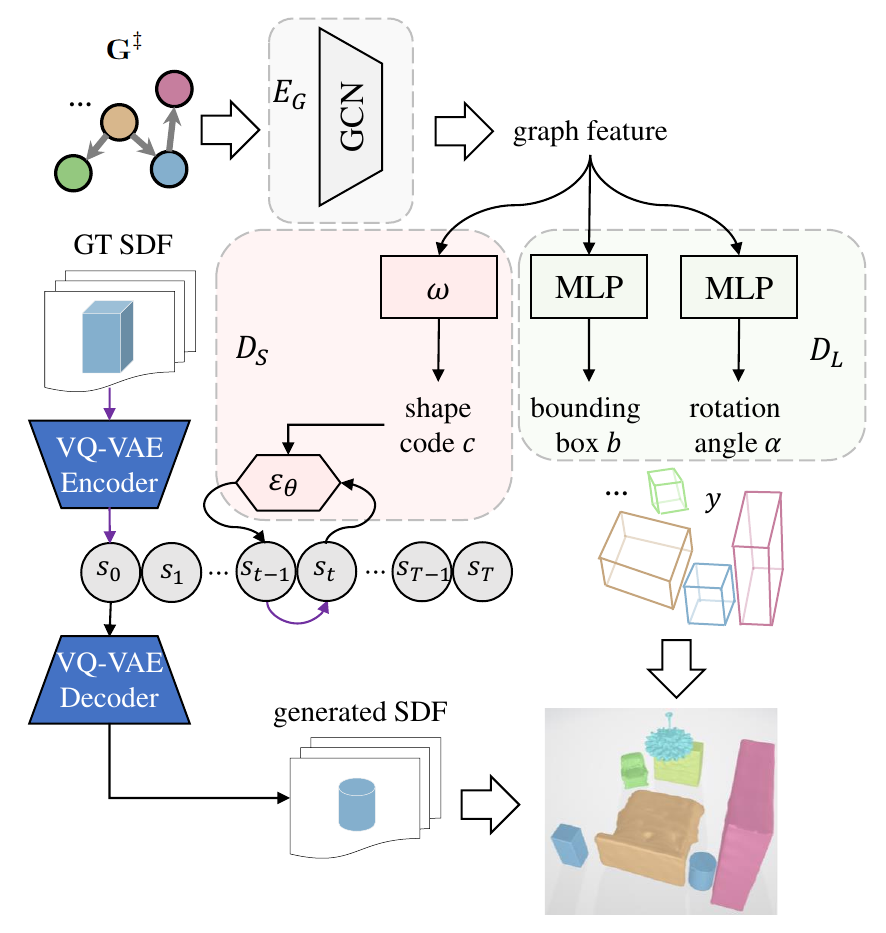}
  \caption{$E_G$ extracts shared graph features for the dual-branch decoder. $D_L$ predicts 6-parameters 3D bounding boxes and their rotation angles. For $D_S$, the latent diffusion model with the denoiser $\epsilon_\theta$ is conditioned on the shape code $c$, and the pre-trained and frozen VQ-VAE is applied to encode GT SDFs into the target latent $s_0$ during training and decode the predicted latent back SDFs during inference.}
  \label{fig:decoder}
\end{figure}

\subsection{Dual-branch Graph-to-Scene Generation}

The graph-to-scene architecture is composed of a graph encoder and a dual-branch scene decoder, shown in Fig. \ref{fig:decoder}.

\vspace{2mm}
\textbf{Graph Encoder} Different from previous approaches that use two separate graph encoders to extract features for guiding the layout and shape decoders independently, we utilize GCNs layers as a unified graph encoder $E_G$ for feature extraction from scene graph representation $\textbf{G}^\ddag$. The shared graph features allow the following two decoders to interact with each other from an early stage. Note that $E_G$ does not share parameters with the GCNs of $\phi$ in the scene graph prior enhancement.

\vspace{2mm}
\textbf{Scene Decoder} The dual-branch decoder consists of a layout branch $D_L$ which produces scene layouts for spatial arrangements and a shape branch $D_S$ which samples 3D object shapes for graph nodes. Specifically, $D_L$ is achieved by two groups of stacked MLP layers in parallel. For the graph node $i$, one group of MLPs is leveraged to predict the 6-parameterized 3D bounding box $b_i$ including width, length, height and centroid coordinates, and another group predicts the angular rotation $\alpha_i$ along the vertical axis. It can be formulated by,
\begin{equation}
  y = \{b\circ \alpha\} = D_L(E_G(\textbf{G}^\ddag)).
  \label{eq:dec_layout}
\end{equation}
In addition, the shape branch $D_S$ is implemented by a group of stacked MLPs (denoted as $\omega$ to distinguish them from the MLPs of the layout branch) and a conditional latent diffusion model. We utilize Truncated SDFs (TSDFs) \cite{curless1996volumetric} with resolution $d_{SDF}$, which can be transformed to 3D meshes, to represent 3D object shapes. The shape codes $c$ is derived from the MLPs,
\begin{equation}
  c = \omega(E_G(\textbf{G}^\ddag)).
  \label{eq:ccc}
\end{equation}
Conditioned on the codes $c$, the latent diffusion model is trained to output latent $s_0$ for 3D shape generation. More specifically, the diffusion model includes a forward diffusion process and a reverse diffusion process. The forward diffusion gradually adds random noise to the target latent $s_0$ using a sequence of increasing noise schedules $\beta_t\in\{\beta_1,...,\beta_{T-1},\beta_T\}$. The shape latent $s_0$ is generated from the GT SDFs by the pre-trained and frozen VQ-VAE \cite{van2017neural} encoder during the training phase. Assuming that $\delta_t:=1-\beta_t$ and $\bar{\delta}_t:=\prod_{\kappa=1}^{t}\delta_\kappa$, the noisy latent $s_t$ is formulated as,
\begin{equation}
  s_t=\sqrt{\bar{\delta}_t}s_0+\sqrt{1-\bar{\delta}_t}\epsilon,
  \label{eq:s_t}
\end{equation}
where $\epsilon$ is a Gaussian noise sampled from ${N} (0,\mathcal{I})$, and the time step $t$ is sampled using the maximum time steps $T$. Starting from $s_T\sim{N} (0,\mathcal{I})$, the reverse diffusion denoises the noisy latent $s_t$ back the target latent $s_0$. To this end, a trainable 3D-UNet \cite{cciccek20163d} is regarded as a denoiser $\epsilon_\theta$ which learns to predict the noise for the reverse diffusion process \cite{ho2020denoising}. Besides, the pre-trained and frozen VQ-VAE decoder is applied to decode the predicted latent back SDFs during the inference phase. 

\subsection{Training Objectives}
The proposed method is trained in an end-to-end manner. 

For scene graph prior enhancement, the training objective is to minimize the Kullback-Liebler (KL) divergence between the posterior distribution $p_\phi(z|x)$ and the distribution $Z$. The KL-based loss is formulated as,
\begin{equation}
  \mathcal{L}_{KL} = D_{KL}(p_\phi(z|x)\Vert p(Z)).
  \label{eq:important}
\end{equation}

For dual-branch graph-to-scene generation, the layout branch and shape branch are trained via layout loss $\mathcal{L}_{layout}$ and shape loss $\mathcal{L}_{shape}$.

On one hand, we employ a commonly-used layout reconstruction loss,
\begin{equation}
  \mathcal{L}_{rec}(\hat{y},y) = \frac{1}{N}\sum_{i=1}^{N}(|\hat{b_i}-b_i|-\sum_{j=1}^{\Lambda}\hat{\alpha}_i^j\log\alpha_i^j),
  \label{eq:rec}
\end{equation}
where $\hat{y} = \{\hat{b}\circ \hat{\alpha}\}$ are the GT layouts. The first term is the 3D bounding box regression loss, and the second is the rotation classification loss. The rotation space is divided into $\Lambda$ bins. However, it is observed that the layout branch trained exclusively with reconstruction loss tends to handle simple scenes but suffers from unrealistic spatial arrangements for complex scenes. To improve the fidelity of generated layouts, we introduce an Intersection-over-Union (IoU) based regularization loss, defined as follows,
\begin{equation}
  \mathcal{L}_{iou}(\hat{y},y) = \sum_{i=1}^{N}(1-IoU(\hat{y}_i, y_i)).
  \label{eq:iou}
\end{equation}
The regularization loss prevents layout collisions by encouraging the similarity between the predicted layouts and the corresponding GT layouts. Thus, the training objective of the layout branch is,
\begin{equation}
  \mathcal{L}_{layout} = \mathcal{L}_{rec} + \eta\mathcal{L}_{iou},
  \label{eq:layout}
\end{equation}
where $\eta$ is used to balance these two loss items.

On the other hand, the shape branch is optimized through training the denoiser $\epsilon_\theta$ with,
\begin{equation}
  \mathcal{L}_{shape} = \Vert\epsilon-\epsilon_\theta(s_t,t,c) \Vert^2,
  \label{eq:shape}
\end{equation}
where $t$ is sampled from discrete values $\{1,...,T-1,T\}$.

The overall training loss function comprises,
\begin{equation}
  \mathcal{L} = \lambda_{KL}\mathcal{L}_{KL} +\lambda_{layout}\mathcal{L}_{layout}+\lambda_{shape}\mathcal{L}_{shape},
  \label{eq:overall}
\end{equation}
where $\lambda_{KL}$, $\lambda_{layout}$ and $\lambda_{shape}$ are weighting factors.

\subsection{Inference}

During the inference phase, the trained model $\phi$ is used to estimate the distribution $Z$. Given a scene graph with initial object and relationship vectors $\mathcal{V}_O$ and $\mathcal{V}_R$, the random vectors $\mathcal{V}_O^Z$ are sampled from the estimated distribution $Z$ for graph nodes $O$. Meanwhile, the pre-trained CLIP text encoder is adopted to generate $\mathcal{V}_O^C$ and $\mathcal{V}_R^C$ for graph nodes $O$ and edges $R$. With aggregated relationship triplets, the pre-trained LLM is utilized to generate a sentence embedding $\mathcal{V}^{F}$. These vectors are incorporated as shown in Equation \eqref{eq:update_graph}, resulting in the graph representation $\textbf{G}^\ddag$.

Next, the trained graph encoder $E_G$ infers graph features from the graph representation $\textbf{G}^\ddag$. The object layouts $y$ and the shape codes $c$ are predicted via Equations \eqref{eq:dec_layout} $\&$ \eqref{eq:ccc}. The latent $s_0'$ is derived using the trained $c$-conditioned denoiser $\epsilon_\theta$ by,
\begin{equation}
  s_0'=\tfrac{1}{\sqrt{\bar{\delta}_t}}(s_t-\sqrt{1-\bar{\delta}_t}\epsilon_\theta(s_t,t,c)),
  \label{eq:s_0}
\end{equation}
and it is then leveraged to reconstruct SDFs with the pre-trained VQ-VAE decoder. Finally, 3D scenes can be synthesized by fitting the 3D object shapes to the generated layouts.

\begin{figure*}[http]
  \centering
  \includegraphics[width=\textwidth]{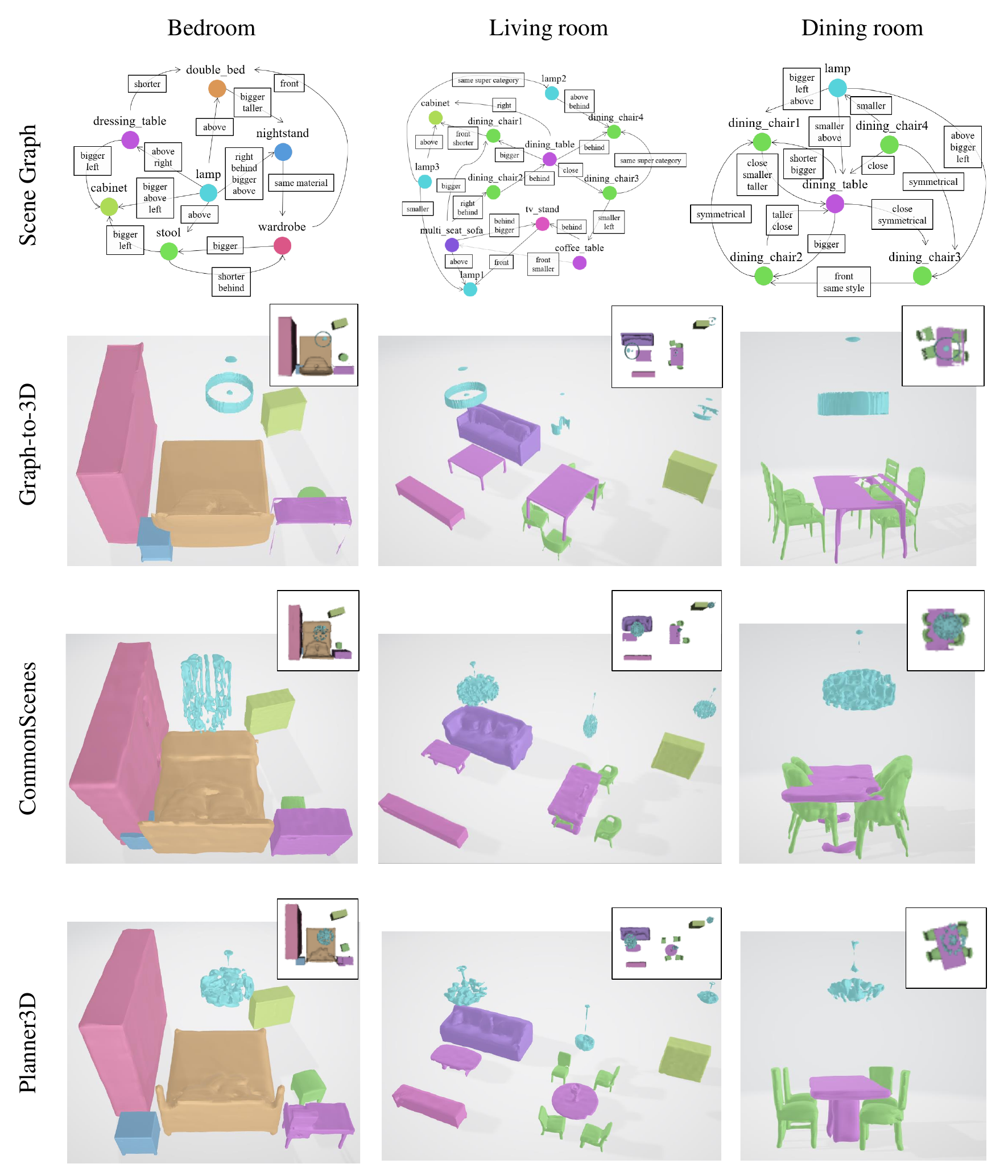}
  \caption{Qualitative examples of bedroom, living room and dining room. Compared to Graph-to-3D \cite{dhamo2021graph} and CommonScenes \cite{zhai2023commonscenes}, Planner3D shows higher fidelity with less interpenetrating phenomena. 
  }
  \label{fig:result}
\end{figure*}

\section{Experiments}

In this section, we clarify the experimental settings, including dataset, evaluation metrics and implementation details. Then, the quantitative and qualitative results are shown and analyzed based on comparisons. Ablations studies are conducted to further verify the effects of the proposed components. Besides, we provide a user study. 

\subsection{Dataset}

Our method is validated on SG-FRONT \cite{zhai2023commonscenes}, a recent benchmark dataset providing scene graph annotations for 3D-FRONT dataset \cite{fu20213d} which populates 3D scenes using furniture objects from 3D-FUTURE dataset \cite{fu20213d2}. SG-FRONT covers three room types of indoor scenes: 4,041 bedrooms, 900 dining rooms, and 813 living rooms. 
Following the original splits \cite{zhai2023commonscenes}, 3,879 bedrooms, 614 dining rooms, and 544 living rooms are used for training, and the remaining ones are used for testing. Each room typically contains multiple objects, ending up with 45K object samples in total.
The scene compositions are described by scene graphs, where nodes are defined by object categories and edges are mainly spatial relationships (e.g., \textit{left/right of}, \textit{close by}) between objects. Overall, SG-FRONT dataset has 35 object categories and 15 relationship categories. 

\subsection{Evaluation Metrics}
The synthesized 3D scenes are evaluated from three aspects:

\vspace{2mm}
\noindent\textbf{Scene graph consistency.} Using pre-defined geometric constraints, the scene graph consistency is measured to assess the spatial consistency between the generated layouts and the corresponding relationships specified in the given graph edges. Specifically, we validate "easy" relationships including \textit{left/right}, \textit{front/behind}, \textit{bigger/smaller than}, \textit{taller/shorter than}, as well as "hard" relationships including \textit{close by} and \textit{symmetrical to}. These relationships correspond to the geometric rules depicted in Table \ref{tab:supp_define}. $IoU(\cdot)$, $Cor(\cdot)$, $Dis(\cdot)$ and $Fli(\cdot)$ denote the processes of calculating IoUs, bounding box corners, distances, and flipping based on $x$-axis or $y$-axis, respectively. The bounding box $b_i$ is composed of the bottom center coordinates $\{x_i,y_i,z_i\}$, width $w_i$, length $l_i$ and height $h_i$. For fair comparisons, we employ the thresholds used in previous works.

\vspace{2mm}
\noindent\textbf{Object-level fidelity.}
The shape fidelity is reported by Chamfer Distance (CD) \cite{fan2017point} based Minimum Matching Distance (MMD), Coverage (COV), and 1-Nearest Neighbor Accuracy (1-NNA) \cite{yang2019pointflow}, which are computed via Equations \eqref{eq:cd}, \eqref{eq:mmd}, \eqref{eq:cov} and \eqref{eq:nna}.

\begin{equation}
  CD(P, \hat{P})=\sum_{p\in P}\mathop{\min}_{\hat{p}\in \hat{P}}\parallel p-\hat{p}\parallel_2^2+\sum_{\hat{p}\in \hat{P}}\mathop{\min}_{p\in P}\parallel p-\hat{p}\parallel_2^2,
  \label{eq:cd}
\end{equation}
where $P$ and $\hat{P}$ are point clouds sampled from the generated shape and the corresponding GT shape, respectively.

\begin{equation}
  MMD(X, \hat{X})=\frac{1}{\left|\hat{X}\right|}\sum_{\hat{P}\in \hat{X}}\mathop{\min}_{P\in X}CD(P,\hat{P}),
  \label{eq:mmd}
\end{equation}

\begin{equation}
  COV(X, \hat{X})=\frac{\left|\{arg min_{\hat{P}\in \hat{X}}CD(P,\hat{P})|P\in X\}\right|}{\left|\hat{X}\right|},
  \label{eq:cov}
\end{equation}

\begin{equation}
    1-NNA(X, \hat{X})=
    \frac{\sum_{P\in X}\mathbb{I}[N_P\in X]+\sum_{\hat{P}\in \hat{X}}\mathbb{I}[N_{\hat{P}}\in \hat{X}]}{\left|X\right|+\left|\hat{X}\right|},
\label{eq:nna}
\end{equation}
where $X$ and $\hat{X}$ are the generated scene collection and the GT scene collection, respectively. In addition, $\mathbb{I}[\cdot]$ is the indicator function, and $N_P$ is the CD-based nearest neighbor of $P$ in $X_{-P}=\hat{X}\cup X-\{P\}$.

\vspace{2mm}
\noindent\textbf{Scene-level fidelity.} To evaluate the quality of scene synthesis from a global perspective, we report two metrics: FID \cite{heusel2017gans} and KID \cite{binkowski2018demystifying}, which are calculated with the top-down renderings of the synthesized and reference 3D scenes. Following prior works, scene-level fidelity acts as the most important evaluation metric.

\begin{table}[http]
    \caption{Definitions of geometric constraints between two instances $i$ and $j$.}
    \label{tab:supp_define}
    \centering
    \begin{tabular}{c|c}
    \hline
    Relationship & Rule \\
    \hline
    \textit{left of} & $x_i < x_j$ and $IoU(b_i, b_j) < 0.3$\\
    \textit{right of} & $x_i > x_j$ and $IoU(b_i, b_j) < 0.3$\\
    \hline
    \textit{front of} & $y_i < y_j$ and $IoU(b_i, b_j) < 0.3$\\
    \textit{behind of} & $y_i > y_j$ and $IoU(b_i, b_j) < 0.3$\\
    \hline
    \textit{bigger than} & $(w_il_ih_i-w_jl_jh_j)/w_il_ih_i > 0.15$\\
    \textit{smaller than} & $(w_il_ih_i-w_jl_jh_j)/w_il_ih_i < -0.15$\\
    \hline
    \textit{taller than} & $((z_i+h_i)-(z_j+h_j))/(z_i+h_i)>0.1$ \\
    \textit{shorter than} & $((z_i+h_i)-(z_j+h_j))/(z_i+h_i)<-0.1$ \\
    \hline
    \textit{close by} & $Dis(Cor(b_i),Cor(b_j))<0.45$\\
    \hline
    \textit{symmetrical to} & $Dis(Fli(b_i),b_j)<0.45$\\
    \hline
    \end{tabular}
\end{table}

\subsection{Implementation Details}
Regarding scene graph prior enhancement, we set the feature dimension $Q$ as 64. The CLIP-based vectors are 512-dimensional, i.e., $\mathcal{V}_O^C \in \mathbb{R}^{\mathrm{N\times512}}$ and $\mathcal{V}_R^C \in \mathbb{R}^{\mathrm{M\times512}}$. Since Instructor Transformer can generate text embedding tailored to diverse domains (e.g., medicine, finance, and science) and tasks (e.g., retrieval, and summarization), we exploit the instruction "Represent the Science document for summarization" to specify the LLM usage goal. The dimension of the LLM-based vector $\mathcal{V}^F$ is 768. Five GCN layers are exploited to extract graph $\textbf{G}^\dag$ features, which are then feed with two MLP modules. The dimension of $\mu_b$ and $\sigma_b$ is 48, and the dimension of $\mu_\alpha$ and $\sigma_\alpha$ is 16. After the concatenation process, the resulting  $\mu$ and $\sigma$ are 64-dimensional and the distribution $Z \sim\mathcal{N} (\mu,\sigma)$ is 64-dimensional Gaussian distribution. The layout feature vectors $\mathcal{V}_O^B$ are only used for training and not needed for inference phase, hence, the inference starts from constructing the scene graph representation $\textbf{G}^\ddag$. Regarding dual-branch graph-to-scene generation, the graph encoder $E_G$ employs 5 GCN layers. With shared graph features, the model $\omega$, consisting of 3 stacked MLP layers, predicts the shape code $c$ which is a 1280-dimensional vector and conditions the denoiser $\epsilon_\theta$. The resolution of SDF is 64, i.e. $SDF \in \mathbb{R}^{\mathrm{64\times64\times64}}$. The shape latent $s_0 \in \mathbb{R}^{\mathrm{16\times16\times16}}$. For the latent diffusion model, the time step $T$ is set to 1000. The layout decoder $D_L$ takes graph features as inputs and utilize two groups of 3 stacked MLPs to estimate 6-parameterized bounding box $b$ and rotation angle $\alpha$. The whole framework is trained in an end-to-end manner with the AdamW optimizer \cite{loshchilov2018decoupled} and the initial learning rate is set to $1e^{-4}$. The weighting factors $\lambda_{KL}, \lambda_{layout}, \lambda_{shape}$ are set to 1.0, while $\eta$ is set to 0.01. For the rotation classification, $\Lambda$ is set to 24. The batch size is 32. All the experiments are conducted on a single NVIDIA A40 GPU with 46GB memory.

\begin{table*}[t]
    \caption{Comparisons  with Graph-to-3D \cite{dhamo2021graph} and CommonScenes \cite{zhai2023commonscenes} on scene graph consistency (higher is better). The best results are shown in \textbf{bold}.}
    \label{tab:SG}
    \centering
    \begin{tabular}{l|cccc|cc|c}
    \hline
    \multirow{2}{*}{Method} & \multicolumn{4}{c|}{Easy} & \multicolumn{2}{c|}{Hard} & \multirow{2}{*}{mSG}\\
    & \textit{left/right$\uparrow$} & \textit{front/behind$\uparrow$} & \textit{big/small$\uparrow$} & \textit{tall/short$\uparrow$} & \textit{close$\uparrow$} & \textit{symmetrical$\uparrow$} \\
    \hline
    Graph-to-3D \cite{dhamo2021graph} ($\ast$) & \textbf{0.98} & 0.99 & 0.96 & 0.93 & 0.74 & 0.50 & 0.87 \\
    CommonScenes \cite{zhai2023commonscenes} ($\ast$) & \textbf{0.98} & \textbf{1.00} & \textbf{0.97} & \textbf{0.96} & \textbf{0.76} & \textbf{0.62} & \textbf{0.90}\\
    \hline
    Graph-to-3D \cite{dhamo2021graph} & \textbf{0.98} & 0.99 & \textbf{0.97} & 0.96 & 0.75 & 0.64 & \textbf{0.90}\\
    CommonScenes \cite{zhai2023commonscenes} & \textbf{0.98} & 0.99 & \textbf{0.97} & 0.96 & 0.76 & \textbf{0.66} & \textbf{0.90}\\
    Planner3D & \textbf{0.98} & \textbf{1.00} & \textbf{0.97} & \textbf{0.97} & \textbf{0.77} & \textbf{0.66} & \textbf{0.90} \\
    \hline
    \end{tabular}
\end{table*}

\begin{table*}[http]
    \caption{Comparisons  with Graph-to-3D \cite{dhamo2021graph} and CommonScenes \cite{zhai2023commonscenes} on object-level fidelity. MMD ($\times 0.01$), COV ($\%$) and 1-NNA ($\%$) are computed. Total is calculated as the mean values across categories. The best results are shown in \textbf{bold}.}
    \label{tab:MMD}
    \centering
    \begin{tabular}{l|ccc|ccc|ccc}
    \hline
    \multirow{2}{*}{Category} & \multicolumn{3}{c|}{MMD$\downarrow$} & \multicolumn{3}{c|}{COV$\uparrow$} & \multicolumn{3}{c}{1-NNA$\downarrow$}\\
    & \cite{dhamo2021graph} & \cite{zhai2023commonscenes} & Ours & \cite{dhamo2021graph} & \cite{zhai2023commonscenes} & Ours & \cite{dhamo2021graph} & \cite{zhai2023commonscenes} & Ours\\
    \hline
    Bed & \textbf{1.27} & 1.47 & 1.73 & \textbf{10.34} & 5.17 & 4.31 & 97.41 & \textbf{92.62} & 93.44 \\
    Nightstand & \textbf{3.00} & 7.12 & 4.49 & \textbf{11.21} & 6.03 & 6.03 & 98.28 & \textbf{93.60} & \textbf{93.60} \\
    Wardrobe & \textbf{1.15} & 3.11 & 1.41 & \textbf{7.76} & 2.59 & 6.03 & 97.41 & 97.56 & \textbf{91.87} \\
    Chair & \textbf{2.28} & 2.57 & 3.10 & 8.62 & \textbf{9.48} & 6.90 & 97.41 & \textbf{93.23} & \textbf{93.23} \\
    Table & 5.68 & 4.98 & \textbf{4.27} & 8.62 & \textbf{12.07} & 10.34 & 96.98 & \textbf{86.57} & 88.06 \\
    Cabinet & 2.57 & 3.45 & \textbf{2.41} & 5.17 & 5.17 & \textbf{9.48} & 96.45 & 96.21 & \textbf{88.64} \\
    Lamp & 4.00 & 2.78 & \textbf{2.44} & 7.76 & 10.34 & \textbf{12.07} & 98.71 & 86.86 & \textbf{81.75} \\
    Shelf & 6.60 & 5.43 & \textbf{3.14} & 6.67 & 20.00 & \textbf{26.67} & 96.00 & \textbf{56.52} & 65.22 \\
    Sofa & 1.31 & 1.11 & \textbf{1.08} & 0.86 & 6.90 & \textbf{9.48} & 99.04 & \textbf{93.75} & 96.88 \\
    TV stand & 0.98 & \textbf{0.62} & 0.84 & 2.59 & \textbf{12.93} & 6.03 & 98.02 & \textbf{90.91} & 94.70\\
    \hline
    Total & 2.88 & 3.26 & \textbf{2.49} & 6.96 & 9.07 & \textbf{9.73} & 97.57 & 88.78 & \textbf{88.74} \\
    \hline
    \end{tabular}
\end{table*}

\vspace{5mm}

\begin{table*}[t]
    \caption{Comparisons with Graph-to-3D \cite{dhamo2021graph} and CommonScenes \cite{zhai2023commonscenes} on scene-level fidelity in SG-FRONT dataset \cite{zhai2023commonscenes} (lower is better). FID and KID ($\times 0.001$) are computed based on the top-down images (256$\times$256 pixels) rendered from the synthesized and reference 3D scenes. ($\ast$) denotes the shape retrieval version of the method where only layout branch is utilized. The best results are shown in \textbf{bold}.}
    \label{tab:FID_KID}
    \centering
    \begin{tabular}{l|cc|cc|cc|cc}
    \hline
    \multirow{2}{*}{Method} & \multicolumn{2}{c|}{Bedroom} & \multicolumn{2}{c|}{Living room} & \multicolumn{2}{c|}{Dining room} & \multicolumn{2}{c}{All}\\
    & FID$\downarrow$ & KID$\downarrow$ & FID$\downarrow$ & KID$\downarrow$ & FID$\downarrow$ & KID$\downarrow$ & FID$\downarrow$ & KID$\downarrow$  \\
    \hline
    Graph-to-3D \cite{dhamo2021graph} ($\ast$) & \textbf{60.48} & \textbf{7.85} & 81.31 & 3.97 & 67.40 & 5.88 & \textbf{45.89} & \textbf{4.34} \\
    CommonScenes \cite{zhai2023commonscenes} ($\ast$) & 64.91 & 13.23 & \textbf{78.24} & \textbf{2.65} & \textbf{65.48} & \textbf{4.13} & 47.22 & 5.98\\
    \hline
    Graph-to-3D \cite{dhamo2021graph} & 68.37 & 17.97 & 92.50 & 18.48 & 71.13 & 7.76 & 51.91 & 9.25\\
    CommonScenes \cite{zhai2023commonscenes} & 65.26 & 12.79 & 85.87 & 8.89 & 69.67 & 6.71 & 49.89 & 7.37\\
    Planner3D & \textbf{60.47} & \textbf{9.30} & \textbf{82.99} & \textbf{8.51} & \textbf{69.40} & \textbf{6.39} & \textbf{47.34} & \textbf{6.13} \\
    \hline
    \end{tabular}
\end{table*}

\vspace{5mm}

\begin{figure*}[tb]
  \centering
  \includegraphics[width=.95\textwidth]{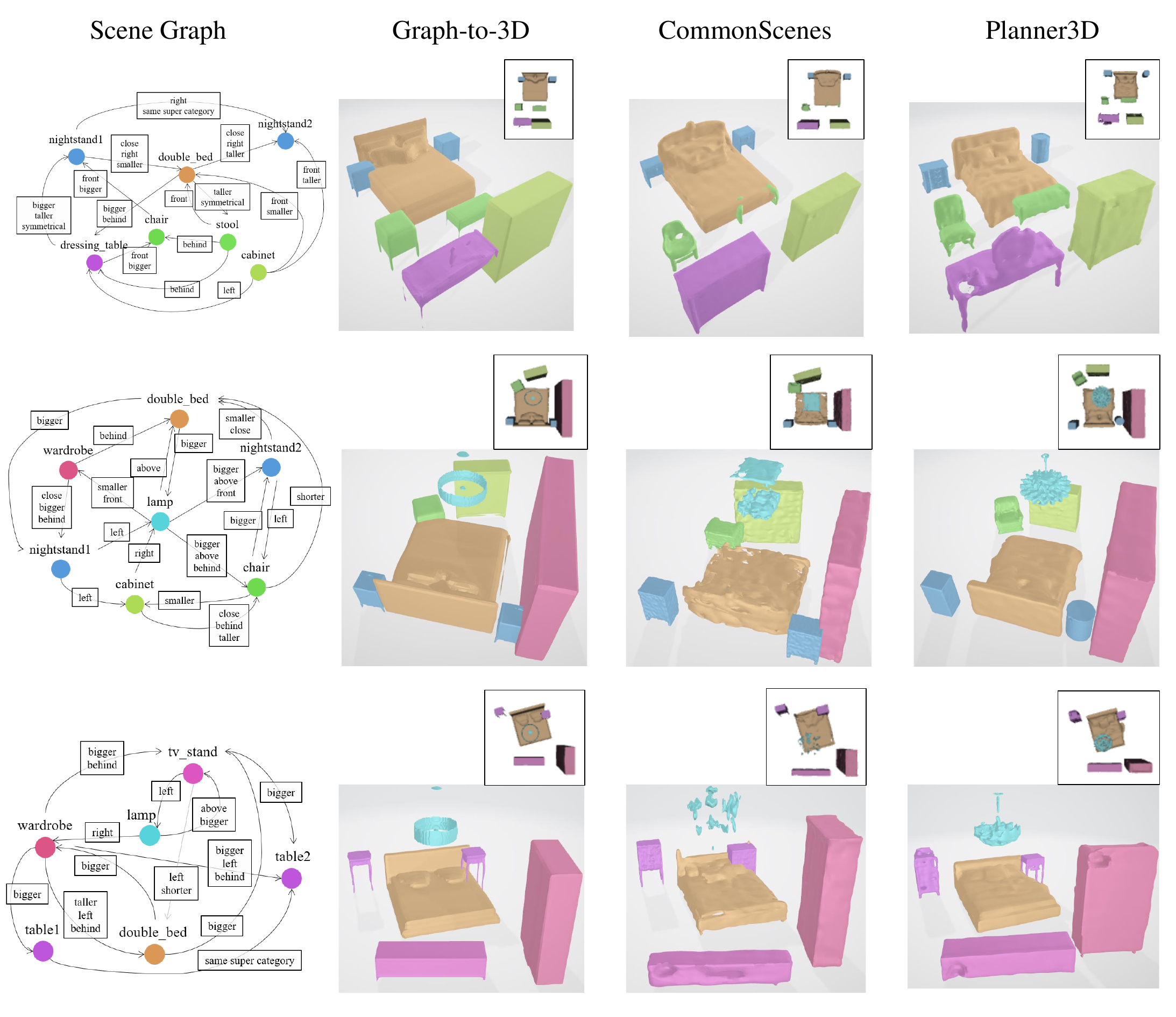}
  \caption{Additional qualitative comparisons on bedroom scenes. A side view of each 3D scene is provided with a top-down view.  
  }
  \label{fig:supp_res_bedroom}
\end{figure*}

\subsection{Quantitative Results and Comparisons}

The proposed Planner3D is compared with two recent state-of-the-art scene graph-to-3D scene synthesis methods: Graph-to-3D \cite{dhamo2021graph} and CommonScenes \cite{zhai2023commonscenes} \footnote{All the results of Graph-to-3D \cite{dhamo2021graph} and CommonScenes \cite{zhai2023commonscenes} presented in this paper are reproduced using the official code with default settings.}.

First, we report the performance in terms of scene graph consistency, as shown in Table \ref{tab:SG} where mSG indicates the mean scene graph consistency computed on both easy and hard relationships. In general, with joint layout-shape learning frameworks, (semi-) generative methods derive better consistency than retrieval-based methods $(\ast)$ that only learn layout synthesis. These methods yield close mSG values due to their comparable performance on easy relationships. Still, Planner3D achieves the best results overall because it performs relatively superior to SOTAs on hard relationships.  

Second, we provide object-level fidelity comparisons of shape generative methods. 
Table \ref{tab:MMD} shows the results of 10 object categories on MMD, COV and 1-NNA. 
In terms of MMD score (lower is better), Planner3D is 0.39 lower than Graph-to-3D \cite{dhamo2021graph} and 0.77 lower than CommonScenes \cite{zhai2023commonscenes}.
In terms of COV score (higher is better), Planner3D is 2.77 higher than Graph-to-3D \cite{dhamo2021graph} and 0.7 higher than CommonScenes \cite{zhai2023commonscenes}.
In terms of 1-NNA score (lower is better), Planner3D is 8.83 lower than Graph-to-3D \cite{dhamo2021graph} and 0.04 lower than CommonScenes \cite{zhai2023commonscenes}.
Overall, our method attains better results compared to Graph-to-3D \cite{dhamo2021graph} and CommonScenes \cite{zhai2023commonscenes}, which suggests the object-level shape generation ability of the proposed method. 

Lastly, we report the results on scene-level fidelity, as shown in Table \ref{tab:FID_KID}. 
The top two rows are shape retrieval-based methods (i.e., only layout branch is trained and tested) and the bottom three are joint shape and layout learning-based methods. 
Since 3D shapes are directly retrieved from the database, the retrieval-based methods $(\ast)$ usually achieve lower FIDs and KIDs than (semi-) generative methods. 
However, shape retrieval-based methods are heavily bounded by the scale and reliability of the given databases. 
For all the room types, our method has 47.34 FID score and 6.13 KID score.
Compared to the Graph-to-3D \cite{dhamo2021graph}, our method has reached 4.57 lower FID score, 3.12 lower KID score.
Compared to the state-of-the-art CommonScenes \cite{zhai2023commonscenes}, our method has reached 2.55 lower FID score, 1.24 lower KID score.
Overall, Planner3D obtains the best FID and KID scores on all the room types (lower score indicates better performance). 
Surprisingly, the proposed method even achieves the best performance with retrieval-based methods on the FID score in Bedroom scenes. Furthermore, Table \ref{tab:5times} reports the FID and KID values on \textbf{all} the room types by running the evaluations 5 times, demonstrating the effectiveness and stability of Planner3D.

\begin{table}[t]
    \caption{Scene-level fidelity comparisons on all room types by running the evaluations 5 times. The best results are shown in \textbf{bold}.}
    \label{tab:5times}
    \centering
    \begin{tabular}{l|cc}
    \hline
    Method & FID$\downarrow$ & KID$\downarrow$ \\
    \hline
    Graph-to-3D \cite{dhamo2021graph} & 51.916$\pm$0.002 & 9.250$\pm$0.004\\
    CommonScenes \cite{zhai2023commonscenes} & 49.859$\pm$0.303 & 7.533$\pm$0.262\\
    Planner3D & \textbf{47.486$\pm$0.278} & \textbf{6.254$\pm$0.394} \\
    \hline
    \end{tabular}
    \vspace{-5mm}
\end{table}


\begin{figure*}[tb]
  \centering
  \includegraphics[width=\textwidth]{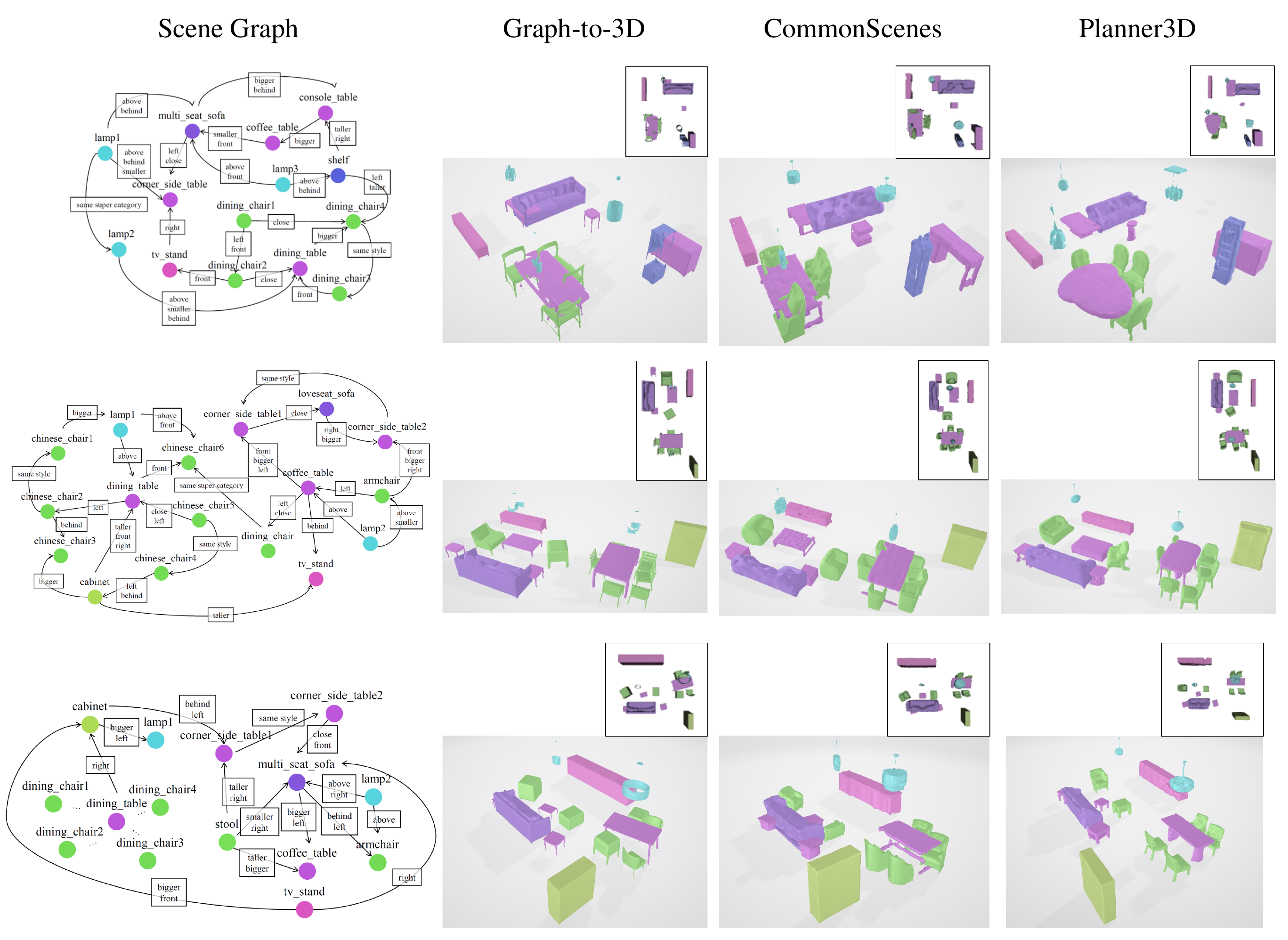}
  \caption{Additional qualitative comparisons on living room scenes. A side view of each 3D scene is provided with a top-down view. 
  }
  \label{fig:supp_res_livingroom}
\end{figure*}

\begin{figure*}[tb]
  \centering
  \includegraphics[width=.9\textwidth]{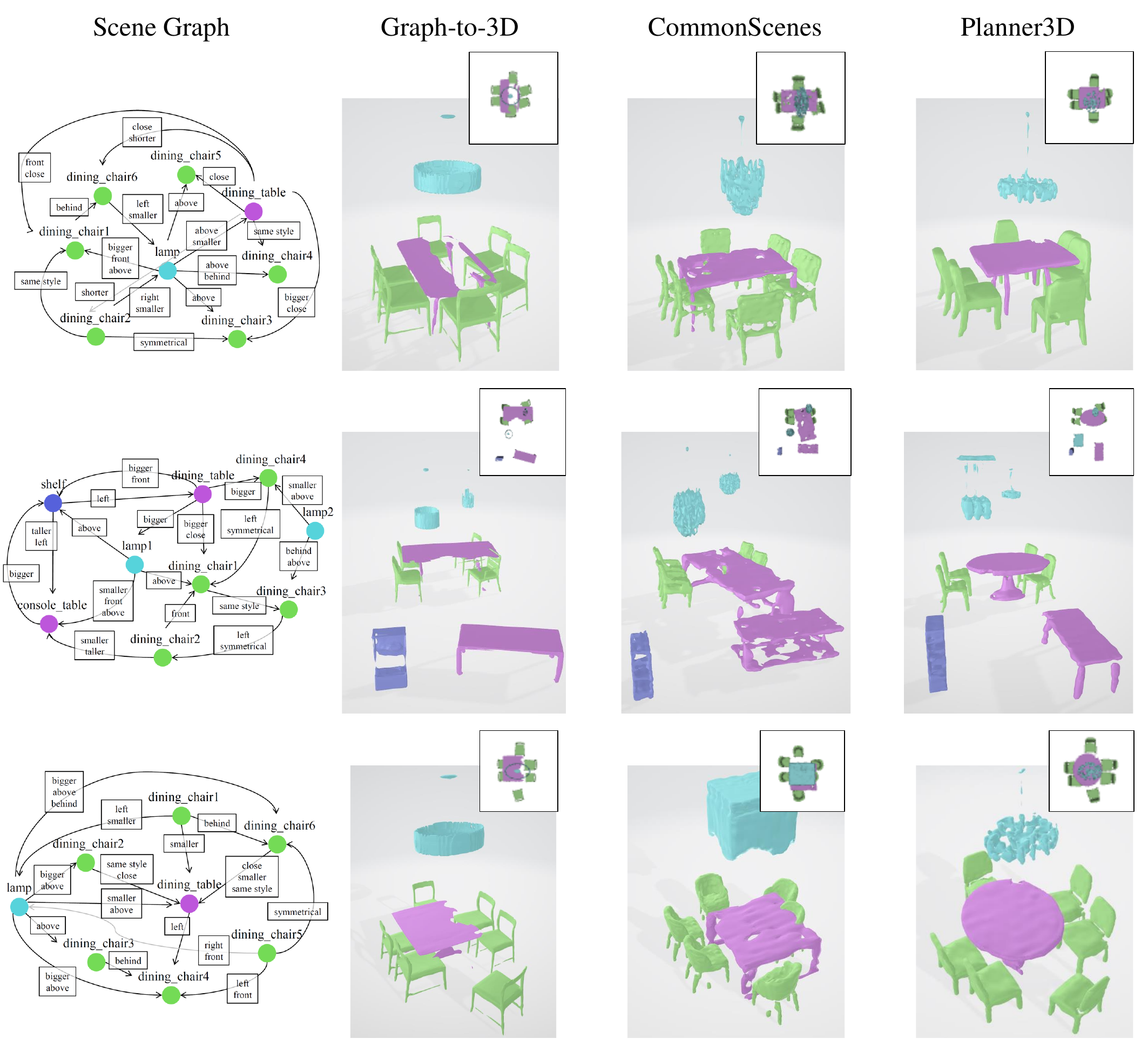}
  \caption{Additional qualitative comparisons on dining room scenes. A side view of each 3D scene is provided with a top-down view. 
  }
  \label{fig:supp_res_diningroom}
\end{figure*}

\subsection{Qualitative Results and Comparisons}
Given the scene graph inputs, qualitative results on bedroom, living room and dining room are shown in Fig. \ref{fig:result}. Overall, Graph-to-3D \cite{dhamo2021graph} and CommonScenes \cite{zhai2023commonscenes} tend to synthesize unrealistic 3D scenes in which ambiguous shapes (e.g., broken dining table) and interpenetrating phenomena (e.g., table and chair intersect with each other) commonly happen. In contrast, Planner3D is able to generate much more realistic 3D scenes with reasonable spatial arrangements and shape generation. 

On one hand, our layout synthesis is constrained with IoU-based regularization, which efficiently alleviates object collision problems compared to those un-regularized layout regression approaches, as shown in the nightstand-double bed-wardrobe case of Fig. \ref{fig:result} bedroom. On the other hand, we notice that the 3D shapes generated from Planner3D are more controllable and not as random as prior works. 
It can be seen from the table-and-chair cases in Fig. \ref{fig:result}, the table is generated as round shape (in living room) or with single table leg (in dining room), which match the arrangements of the surrounding chairs. 
Without special designs, our shape branch benefits from the proposed scene graph prior enhancement mechanism, and the shared graph feature encoding with the layout branch which enables joint layout-shape learning at an early stage.

More visualizations of the bedroom, living room and dining room results are provided in Figs. \ref{fig:supp_res_bedroom}, \ref{fig:supp_res_livingroom}, and \ref{fig:supp_res_diningroom}, respectively. It can be seen that Graph-to-3D \cite{dhamo2021graph} usually generates 3D scenes composed of regular object shapes and performs well for simple scenes such as bedrooms. However, the quality of generated objects degrades as the number of objects increases, e.g., on living rooms and dining rooms. Note that due to limited space, we partially visualize the scene graphs, e.g., the last scene graph of Fig. \ref{fig:supp_res_livingroom}. Besides, it is observed that Graph-to-3D suffers from limited shape diversity. For instance, it generates a specific 3D shape for the object \textit{lamp} across different scenes. Benefiting from the diffusion-based shape branch, CommonScenes \cite{zhai2023commonscenes} is able to synthesize diverse and plausible 3D object shapes, but the generated scenes are often unrealistic due to low-quality shapes and object collisions. In contrast, Planner3D performs better at object-level and scene-level. It is worth pointing out that the 3D scenes generated by Planner3D exhibit more realistic shapes and reasonable layout configurations.

\begin{table*}[t]
    \caption{Ablation study results on all room types of SG-FRONT dataset. The best results are shown in \textbf{bold}.}
    \label{tab:ablation}
    \centering
    \begin{tabular}{l|cc|ccccccc}
    \hline
    \multirow{2}{*}{Method} & \multicolumn{2}{c|}{Scene-level fidelity} & \multicolumn{7}{c}{Scene graph consistency}\\
    & FID$\downarrow$ & KID$\downarrow$ & mSG$\uparrow$ & \textit{left/right$\uparrow$} & \textit{front/behind$\uparrow$} & \textit{big/small$\uparrow$} & \textit{tall/short$\uparrow$} & \textit{close}$\uparrow$ & \textit{symmetrical}$\uparrow$ \\
    \hline
    Planner3D (Ours) & \textbf{47.34} & \textbf{6.13} & \textbf{0.90} & \textbf{0.98} & \textbf{1.00} & \textbf{0.97} & \textbf{0.97} & \textbf{0.77} & \textbf{0.66}\\
    \quad w/o \textit{LLM} & 47.71 & 7.94 & 0.88 & \textbf{0.98} & 0.99 & 0.96 & 0.95 & \textbf{0.77} & 0.54 \\
    \quad w/o \textit{Reg} & 48.15 & 6.93 & \textbf{0.90} & \textbf{0.98} & 0.99 & \textbf{0.97} & \textbf{0.97} & 0.76 & 0.62\\
    \hline
    \end{tabular}
\end{table*}

\begin{table}[t]
    \caption{Ablation study results w.r.t. layout discriminator and unified graph encoder. The mSG values are computed on both easy and hard relationships. The best results are shown in \textbf{bold}.}
    \label{tab:ablation_dis}
    \centering
    \begin{tabular}{l|ccc}
    \hline
    Method & FID$\downarrow$ & KID$\downarrow$ & mSG$\uparrow$ \\
    \hline
    CommonScenes \cite{zhai2023commonscenes} & 49.89 & 7.37 & 0.90\\
    \quad w/o \textit{$Dis$} & 48.20 & 6.73 & 0.88\\
    \quad w \textit{Uni} & 47.69 & 6.54 & 0.89 \\
    \hline
    Planner3D (Ours) & \textbf{47.34} & \textbf{6.13} & 0.90\\
    \quad w \textit{$Dis$} & 49.09 & 8.74 & \textbf{0.91}\\
    \hline
    \end{tabular}
\end{table}

\subsection{Ablation Study}
\begin{figure}[tb]
  \centering
  \includegraphics[width=\columnwidth]{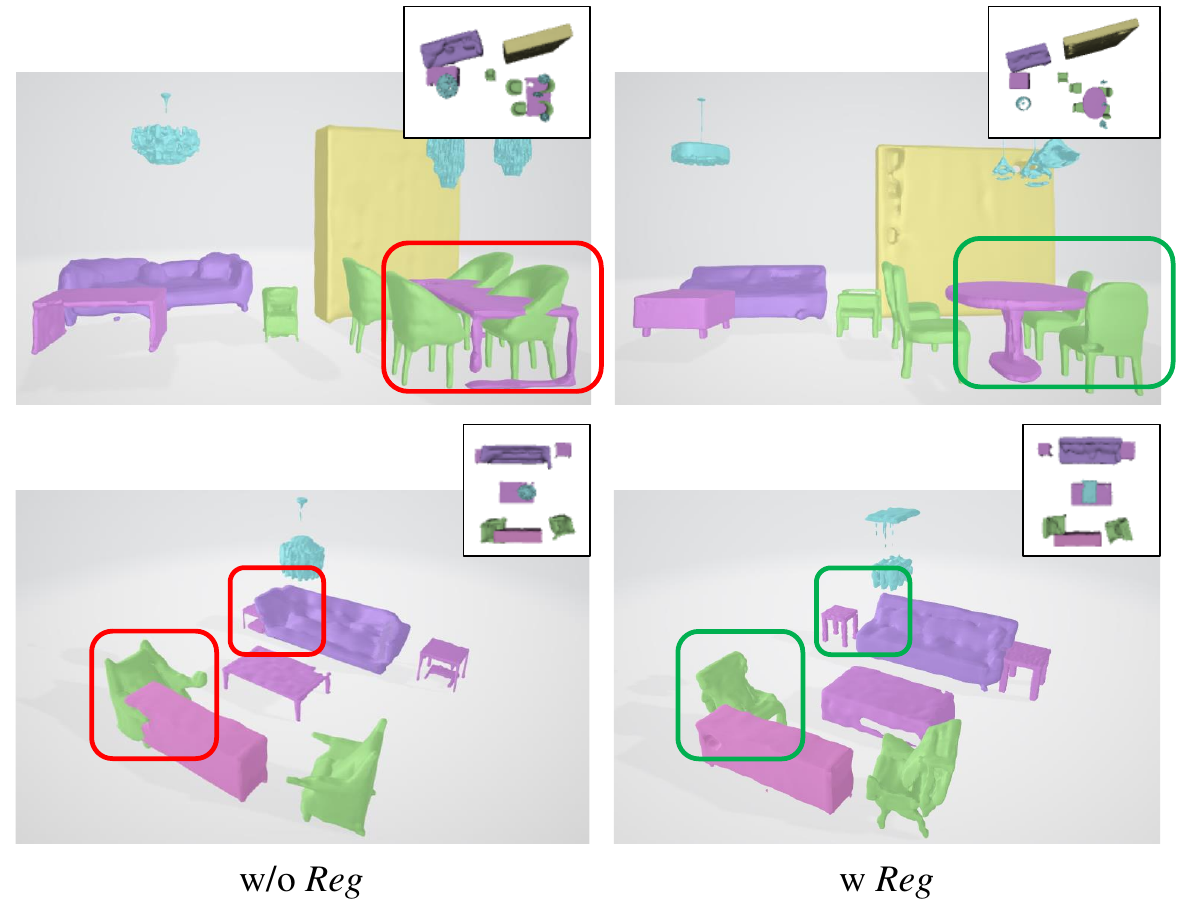}
  \caption{Without \textit{Reg}, unrealistic layouts (highlighted in red rectangles) frequently appear. In contrast, applying \textit{Reg} typically results in well-structured layouts, highlighted in green rectangles. A side view of each 3D scene is provided with a top-down view.
  }
  \label{fig:abla}
\end{figure}

Firstly, we access the effects of the proposed key components in Planner3D: LLM-based scene graph prior enhancement (abbreviated as \textit{LLM}) and explicit layout regularization (abbreviated as \textit{Reg}). Using the metrics in terms of scene-level fidelity and scene graph consistency, the results are shown in Table \ref{tab:ablation}. The first row illustrates our full model, i.e., Planner3D. In contrast, \textit{LLM} supports FID and mSG, and introduces significant improvement on KID and the hard relationship \textit{symmetrical}. This demonstrates the advantage of LLM-based scene graph prior enhancement, which provides rich signals for the following encoding stage. As shown in the last row, \textit{Reg} contributes scene-level fidelity, especially on FID. By jointly applying layout regression and explicit regularization, the spatial arrangements of objects become more realistic with fewer collision issues. Fig. \ref{fig:abla} demonstrates two living room examples which are synthesized without and with \textit{Reg}, respectively. It can be seen from the left column that the scene quality drops significantly due to the overlapping objects. Instead, explicit regularization facilitates reasonable spatial arrangement, as shown in the right column.

Secondly, we investigate the performance of the discriminator (abbreviated as \textit{Dis}) that is commonly used for layout synthesis in prior works \cite{dhamo2021graph, zhai2023commonscenes}, as well as the proposed unified graph encoder (abbreviated as \textit{Uni}). The results are listed in Table \ref{tab:ablation_dis}. Taking the results of the baseline CommonScenes \cite{zhai2023commonscenes} as an example, the FID, KID and mSG values all decrease without \textit{Dis}, which indicates that adversarial training is not an optimal solution to layout synthesis. As can be seen from the bottom two rows, \textit{Dis} offers only slight improvements to scene graph consistency while compromising scene-level fidelity. This proves the claim that GAN-based layout synthesis suffers from training instability and affects the training quality of layout decoder. Rather than training with \textit{Dis}, Planner3D is more robust by explicit layout regularization, achieving better trade-off between scene-level fidelity and scene graph consistency. In addition, even with less learnable parameters, CommonScenes with the unified graph encoder outperforms the raw CommonScenes which utilizes separate graph encoders. The unified graph encoder, which acts as a bridge between LLM-enhanced scene graph prior and dual-branch scene decoder, enables effective integration of layout branch and shape branch under the guidance of shared graph features.

\subsection{User Study}
\label{sec:user}

\begin{figure}[tb]
  \centering
  \includegraphics[width=\columnwidth]{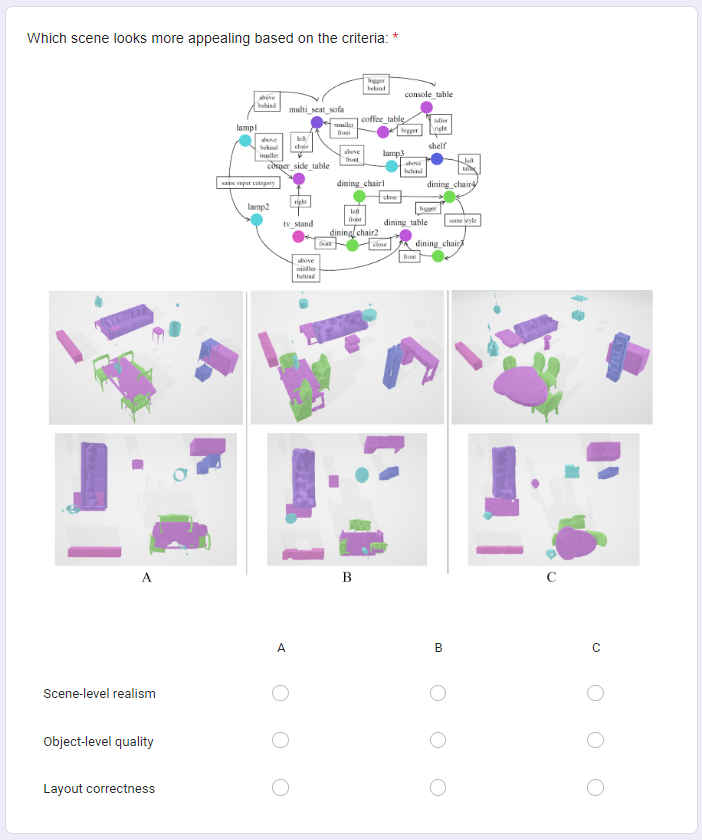}
  \caption{User interface for the user study. 
  }
  \label{fig:supp_interface}
\end{figure}

Although the evaluation metrics can indicate the quality of the generated 3D scenes, they cannot be fully equated with human perception. Therefore, we invited $\sim$30 participants from various professional backgrounds to perform a user study. Fig. \ref{fig:supp_interface} shows the user interface of this study. 

Overall, we randomly sampled 15 scenes covering all room types, and the synthesized 3D scenes were the results of Graph-to-3D \cite{dhamo2021graph}, CommonScenes \cite{zhai2023commonscenes} and our Planner3D, respectively. To ensure the visibility, the scenes were rendered from a side view and a top-down view. Due to space limitation, only parts of relationships (i.e., edges) are shown in the scene graphs. Given the paired scene graph input and the 3D scenes synthesized by different methods, participants were requested to vote for the best scene based on three criteria: scene-level realism (SR), object-level quality (OQ), and layout correctness (LC). They were described to the participants by:

\noindent\textbf{SR}: \textit{Does the scene look realistic and stylistically coherent?}

\noindent\textbf{OQ}: \textit{Do the objects have high-quality 3D shapes?}

\noindent\textbf{LC}: \textit{Does the spatial arrangement of objects appear correct and free of collision?}

Basically, SR requires participants to check whether the scene is realistic and coherent, acting as a global indicator that considers both 3D shapes and layouts. OQ emphasizes the completeness and fidelity of the generated object shapes, while LC denotes the spatial arrangement levels of objects. 

The average results are given in Table \ref{tab:supp_study}. Most of the participants (48.3$\%$) suggested that our generated scenes were preferred over the other methods on the scene-level realism. Where Graph-to-3D was instead preferred 42.5$\%$ and CommonScenes was preferred 9.2$\%$. Regarding the object-level quality, most of the participants (49.7$\%$) preferred the results of Graph-to-3D, without being informed that the inflexibility of its shape generation which relies on pre-trained shape codes from category-wise auto-decoders. CommonScenes and Planner3D were preferred 10$\%$ and 40.3$\%$, respectively. Furthermore, Planner3D was preferred 60.5$\%$ on the layout correctness, much higher than other methods Graph-to-3D 31.7$\%$ and CommonScenes 7.8$\%$, demonstrating its superiority on spatial arrangements of objects. Interestingly, our baseline method CommonScenes obtained the lowest voting rates on all the criteria of this study, even it outperformed Graph-to-3D on the evaluation metrics shown in the main paper. This perceptual user study shows that the proposed method improves compositional 3D scene synthesis compared to the baselines, and enjoys high flexibility without being dependent on per-category shape codes and adversarial training.

\begin{table}[t]
    \caption{User study results via the percentage of voters.}
    \label{tab:supp_study}
    \centering
    \begin{tabular}{l|ccc}
    \hline
    Method & SR & OQ & LC \\
    \hline
    Graph-to-3D \cite{dhamo2021graph} & 42.5$\%$ & 49.7$\%$ & 31.7$\%$\\
    CommonScenes \cite{zhai2023commonscenes} & 9.2$\%$ & 10.0$\%$ & 7.8$\%$\\
    Planner3D & 48.3$\%$ & 40.3$\%$ & 60.5$\%$\\
    \hline
    \end{tabular}
\end{table}

\section{Discussion}

\subsection{Broader Impact}
We believe Planner3D will be applied to interior design, game development, augmented and virtual reality applications. By designing scene graphs, users can create realistic and coherent 3D indoor scenes tailored to their specifications, without the need for specialized 3D modeling expertise. For textured 3D indoor scenes, Planner3D can work with recent 3D scene texturing works, such as MVDiffusion \cite{Tang2023mvdiffusion} and SceneTex \cite{chen2024scenetex}, through providing high-quality semantic scenes. As large language models continue to advance rapidly, the integration of more powerful LLMs, e.g., GPT series \cite{ouyang2022training, OpenAI2023GPT} and Llama \cite{touvron2023llama}, into our pipeline offers substantial potential for improving both the controllability and scalability of 3D scene synthesis.

\subsection{Limitations}
Despite the advancements, there are still some limitations of the proposed method. First of all, the scale of the 3D scene dataset we employed remains small. SG-FRONT \cite{zhai2023commonscenes} offers thousands of samples covering three room types, whereas 3D object datasets \cite{deitke2023objaverse, deitke2024objaverse} typically provide millions of or even more samples. Hence, expanding the existing scene dataset via leveraging large-scale 3D object datasets would be an interesting research direction. The scene graph annotations can be derived from natural language descriptions which are easily obtained. Planner3D is originally developed with a focus on indoor scene synthesis from scene graph. However, the presented pipeline shows potential for adapting to larger-scale outdoor scenes (e.g., \cite{chen2023scenedreamer, lee2024semcity}), which is also a critical and promising topic. On one hand, the open-vocabulary nature and reasoning ability of LLMs could support our scene graph prior enhancement for handling with complex scene-object relationships. On the other hand, our explicit layout regularization can alleviate the limitations of prior works on object spatial arrangements. We will explore these ideas in future work. 

\section{Conclusions}

In this work, we present Planner3D, a simple yet effective scene graph-to-3D scene synthesis method which outperforms previous state-of-the-art methods by a large margin. LLMs are integrated with CLIP to enhance the scene graph representations, enabling a hierarchical understanding of graph priors. A unified graph encoder extracts shared graph features that guide a dual-branch scene decoder. A latent diffusion model is trained at the shape branch. The layout branch is optimized with explicit regularization, instead of adopting adversarial training scheme that suffers from ambiguous spatial arrangements. Experimental results show the superiority of Planner3D in terms of scene graph consistency, object-level fidelity and scene-level fidelity.

\vfill

\end{document}